\documentclass{article}

\usepackage{PRIMEarxiv}

\usepackage[utf8]{inputenc} 
\usepackage[T1]{fontenc}    
\usepackage{hyperref}       
\usepackage{url}            
\usepackage{booktabs}       
\usepackage{amsfonts}       
\usepackage{nicefrac}       
\usepackage{microtype}      
\usepackage{lipsum}
\usepackage{fancyhdr}       
\usepackage{graphicx}       
\graphicspath{{media/}}     

\title{ImageLab: Simplifying Image Processing Exploration for Novices and Experts Alike
\thanks{\textit{\underline{Citation}}: 
\textbf{Authors. Title. Pages.... DOI:000000/11111.}} 
}

\author{
  Sahan Dissanayaka, Oshan Mudanayaka \\
  University of Colombo School of Colombo \\
  University of Colombo \\
  Colombo, Sri Lanka\\
  \texttt{\{tsahandisaai, oshan.ivantha\}@gmail.com} \\
   \And
  Thilina Halloluwa \\
  School of Computer Science\\
  University of Sydney \\
  New South Wales, Sydney\\
  \texttt{thilina.halloluwa@sydney.edu.au} \\
  \AND
  Chameera De Silva \\
  University of Colombo School of Computing \\
  University of Colombo \\
  Colombo, Sri Lanka\\
  \texttt{info.chameera.de@gmail.com} \\
}

\begin{document}
\maketitle

\begin{abstract}
Image processing holds immense potential for societal benefit, yet its full potential is often accessible only to tech-savvy experts. Bridging this knowledge gap and providing accessible tools for users of all backgrounds remains an unexplored frontier. This paper introduces "ImageLab," a novel tool designed to democratize image processing, catering to both novices and experts by prioritizing interactive learning over theoretical complexity. ImageLab not only serves as a valuable educational resource but also offers a practical testing environment for seasoned practitioners. Through a comprehensive evaluation of ImageLab's features, we demonstrate its effectiveness through a user study done for a focused group of school children and university students which enables us to get positive feedback on the tool. Our work represents a significant stride toward enhancing image processing education and practice, making it more inclusive and approachable for all.
\end{abstract}

\keywords{Image Processing, Computer Graphic, Human-Computer Interaction, Usability}

\section{Introduction}

Image processing stands as a pivotal domain in contemporary computing, riding on the coattails of advancements in computer vision and cognitive sciences. Its core revolves around the manipulation and analysis of digital images to both amplify quality and extract pertinent data. Fields as diverse as medicine, agriculture, transportation, and supply chain are beneficiaries of image processing innovations. At its heart, an image is conceptualized as a 2D matrix populated by pixels, and it's atop these individual pixels that image operators exert their influence. The transformations at the pixel level pave the way for discerning patterns and gleaning essential information. Notably, image processing operations often adopt a pipeline paradigm: the output of one operator becomes the input for the subsequent one, lending itself to a linear processing flow. A strong underpinning in linear algebra, which characterizes many image-processing operators, facilitates mathematical representations of images post-transformation.

Despite the prominence of image processing, newcomers to the field often grapple with its foundational concepts, primarily due to a dearth of novice-friendly learning resources. An effective pedagogical approach, especially for beginners, is to immerse them in experiential learning: letting them explore the 'how', ignite their curiosity, and subsequently delve into the intricacies. This methodology echoes the principles of reinforcement learning, where learners assimilate knowledge by iterating through errors, eventually attaining a comprehensive understanding of the subject matter. Such an approach resonates with many students, as it facilitates a tangible engagement with the material, enabling them to internalize core concepts through hands-on experience.

Leveraging core pedagogical insights within the domain of image processing, enhanced by the capabilities of programming and computer science, this study introduces a novel tool named Imagelab (https://github.com/scorelab/imagelab). Initially designed as an open-source project facilitates an intuitive and interactive exploration of image processing concepts in a standalone manner, lowering barriers for newcomers. It not only empowers and excites novices to delve into the intricacies of image processing but also provides seasoned users with a sandboxed environment to experiment before diving into full-fledged development or implementation.

\section{Related Work}
\label{sec:related_works}

Image processing has been a prominent topic for decades, encompassing various categories such as image representation, enhancement, restoration, analysis, and data compression \cite{Sohi2000}. Over the years, there has been substantial growth in methods and frameworks for image processing techniques like segmentation and conversion. These techniques find applications in diverse fields like healthcare, agriculture, and security. The concept of image processing has a history of 40 years, with significant advancements over time. \cite{Cheng2001} presented a survey on colour image segmentation, highlighting its importance in image processing and pattern recognition. The survey discussed techniques like linear transformations, edge detection, and neural networks. The applications of image processing have evolved, from simple image enhancement \cite{Kovasznay1955} to advanced applications like cancer detection \cite{Jain2015}. The teaching of image processing is math-intensive, making curriculum development challenging. Modern frameworks focus on combining image processing with other technologies for enhanced learning and application.

\subsection{Applications and Frameworks for Teaching Image Processing}
\label{applications-and-frameworks-for-teaching-image-processing}

\begin{figure}
\centering
    \includegraphics[width=4.15278in,height=1.69306in]{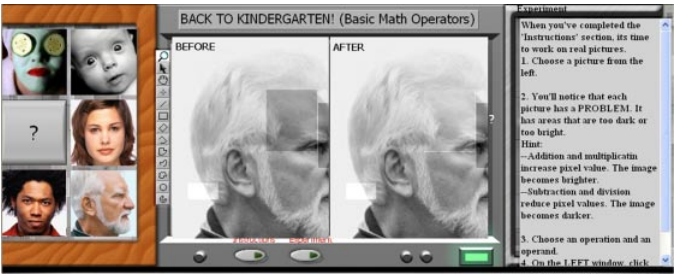}
    \caption{The basic operator experiment program interface \cite{Yapp2008}}
    \label{fig:basic_operator_exp}
\end{figure}

Image processing is gaining traction in fields like engineering, computer science, robotics, machine vision, and mathematics \cite{Yapp2008} as in Figure \ref{fig:basic_operator_exp}. \cite{Yapp2008} introduced an interactive program for teaching digital image processing, covering topics such as basic operators, Fourier transform, noise removal, histogram formations, thresholding, edge detection, and morphological operations. Their curriculum, developed in LabVIEW, offers both detailed pixel-level calculations and real-time filter applications on full-sized images. On the other hand, Ng \cite{Ng1997} emphasizes the mathematical intensity of image processing courses and introduces a computer-based teaching approach using multimedia technology to engage students. The course focuses on using computers to emulate human visual capabilities and process images. COMPEL (Asymetrix Corporation) is used for lectures, and MATROX VISION captures images for the course.

\begin{figure}
    \centering
    \includegraphics[width=.60\textwidth]{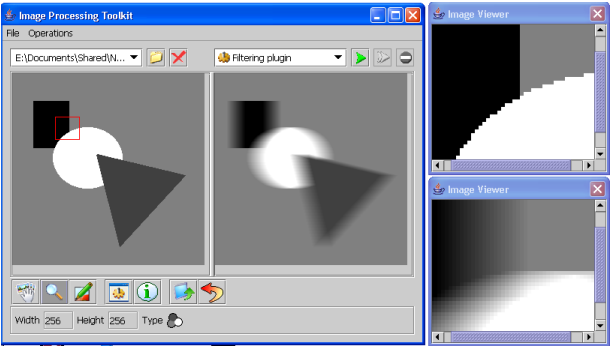}
    \caption{The view on Image Processing Toolkit \cite{Ageenko2005}, illustrating the main application window and enlarged image view }
    \label{fig:img_proc_toolkit_ageenko}
\end{figure}

\cite{Rasure2002} introduced "Khoros," a scientific software environment for teaching and learning image processing. They envisioned the future of education to be dominated by interactive multimedia and collaboration technologies, bridging geographical and cultural gaps, a vision that aligns with the reality of 2023. Ageenko \& la Russa \cite{Ageenko2005} highlighted the challenges of teaching Image Processing due to its abstract nature and emphasized the importance of hands-on experiments for students. They introduced the Image Processing Toolkit (IPT) as in Figure \ref{fig:img_proc_toolkit_ageenko}, a Java-based tool that's platform-independent, open for third-party plugin development, and user-friendly. This tool aims to cater to the growing demand for specialists in the imaging field. \cite{Sohi2000} discussed the significance of "Image Processing" in the tech-driven world and described their research on implementing image processing techniques using Advanced Visual System Express (AVS Express). AVS Express visualizes the program's formation through modules, showing the flow of information in implemented algorithms.

Java has been a significant tool in teaching image processing to students and professionals. \cite{Sage2003} emphasized the student-friendly nature of their system built around ImageJ, a free software for image analysis. They encourage students to start with standard image processing (IP) algorithms in Java and then progressively extend them, facilitating plugin development for ImageJ. In another work, Sage \& Unser \cite{Sage2001} used ImageJ to complement a two-semester course, challenging students with practical imaging problems and teaching them to program standard IP algorithms in Java. \cite{POWELL2001} introduced the Java Vision Toolkit (JVT) for teaching image processing, offering over 50 image operations in a user-friendly GUI. The toolkit, written in Java, is designed for extensibility and includes assignments on thresholding, histogram modification, edge detection, and Hough transform.

However, \cite{Trussell2000} noted that many courses focus on algorithms but neglect the essential display methods needed to showcase algorithm effectiveness. They discussed mathematical functions like Isometric Plots and Contour Plots and emphasized consistent image scaling for accurate comparisons. \cite{Campbell2001} highlighted the shift in student backgrounds, with many now more adept at computer programming than applied mathematics. They introduced DataLab-J, a software laboratory encompassing signal processing, image processing, statistics, and data mining. 

\begin{figure}
    \centering
    \includegraphics[width=4.15278in,height=2.69306in]{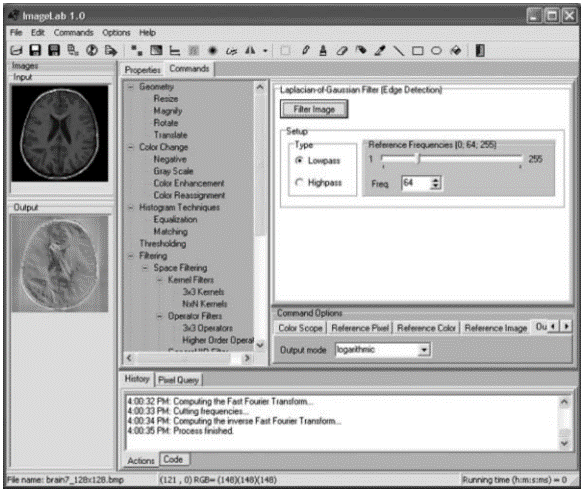}
    \caption{Ayala et al. \cite{Ayala2009} ImageLab's main window showing an MRI image and the result of executing the edge detection command for ventricle segmentation }
    \label{fig:imagelab_ayala_2009}
\end{figure}

\cite{Ayala2009} addressed the challenge faced by researchers unfamiliar with image processing. They proposed a user-friendly interface, ImageLab 1.0, it further depicted in Figure \ref{fig:imagelab_ayala_2009} to simplify common image processing tasks, allowing users to understand and perform operations primarily through mouse clicks while also providing direct access to the software code for enhancements.

Teaching image processing concepts in junior high school is another topic of discussion and research on this is presented by \cite{Barak2012} stating that the overarching purpose of the study was that of integrating the learning of subjects in STEM and linking the learning of these subjects to the children's world and to the digital culture characterization society today. The findings of this research indicated that boys' and girls' achievements were similar throughout the course, and all managed to handle the mathematical knowledge without any particular difficulties, while learners' motivation to engage in the subject was high in the project-based learning part of the course than the theory part \cite{Barak2012}. In an engineering educational context, a Digital Image processing undergraduate course is complex for students due to its multidisciplinary content, and the educational software traditionally used for undergraduate education in digital image processing usually works with good speed but in an individualized manner, \cite{Garcia2015}, presents a web-based tool for teaching of FPGA -- based digital image processing in undergraduate courses and presents a collaboration based solution for teaching image processing at undergraduate level efficiently. The hardware circuit and experimentation workspace proposed by \cite{Garcia2015} is given in diagram below in Figure \ref{fig:img_proc_garcia_2015}.

\begin{figure}
    \centering
    \includegraphics[width=.60\textwidth]{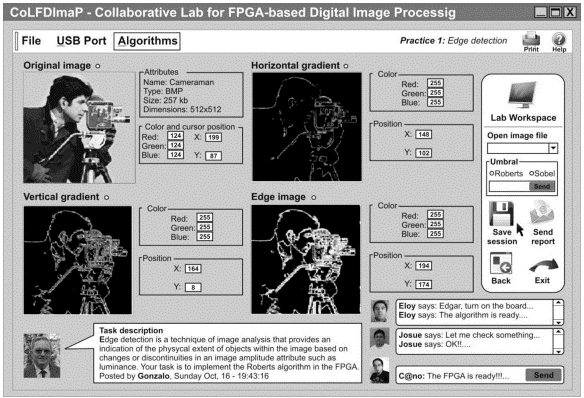}
    \caption{The experimentation workspace of CoLFDlmaP \cite{Garcia2015}}
    \label{fig:img_proc_garcia_2015}
\end{figure}

\cite{silva2015supportive} argue that the teaching of digital image processing (DIP) is often hampered by the intricacy of implementing various techniques, making it challenging for students to visualize DIP results during courses. They introduced a supportive environment for DIP teaching and learning, aiming to facilitate quick visualization and experimentation of methods without the need for software installation or prior programming knowledge. In a more recent study, \cite{jimenez2016teaching} shared their experience using the Project Based Learning (PBL) methodology to teach an undergraduate image processing course at Universidad de los Llanos. Their approach resulted in student projects that applied core concepts of image processing and pattern analysis, underpinned by the PBL methodology. The teaching of image processing is diverse, utilizing various software and frameworks, and is increasingly being integrated with other technologies.

\subsection{Human Computer Interaction and Image Processing}

Creating tailored image processing applications demands significant time and expertise, often hindering its adoption in industries despite growing demands across various sectors like medicine, security, and biotechnology. \cite{Clouard2011} introduced an interactive system designed to generate image processing software. Their design revolves around two core human-computer interaction models: the formulation model, which organizes necessary information for application development, and the interaction model, which outlines how to gather this information from users. Beyond this, numerous applications merge image processing and human-computer interaction, including measuring apple sweetness and predicting fruit diseases \cite{Kumar2022}, analyzing dance movements \cite{Jin2022}, and facilitating LED-to-Camera communication by analyzing recorded images or videos \cite{Dalgic2022}.

\section{System Design and Implementation}

\begin{figure}
    \centering
    \includegraphics[width=.80\textwidth]{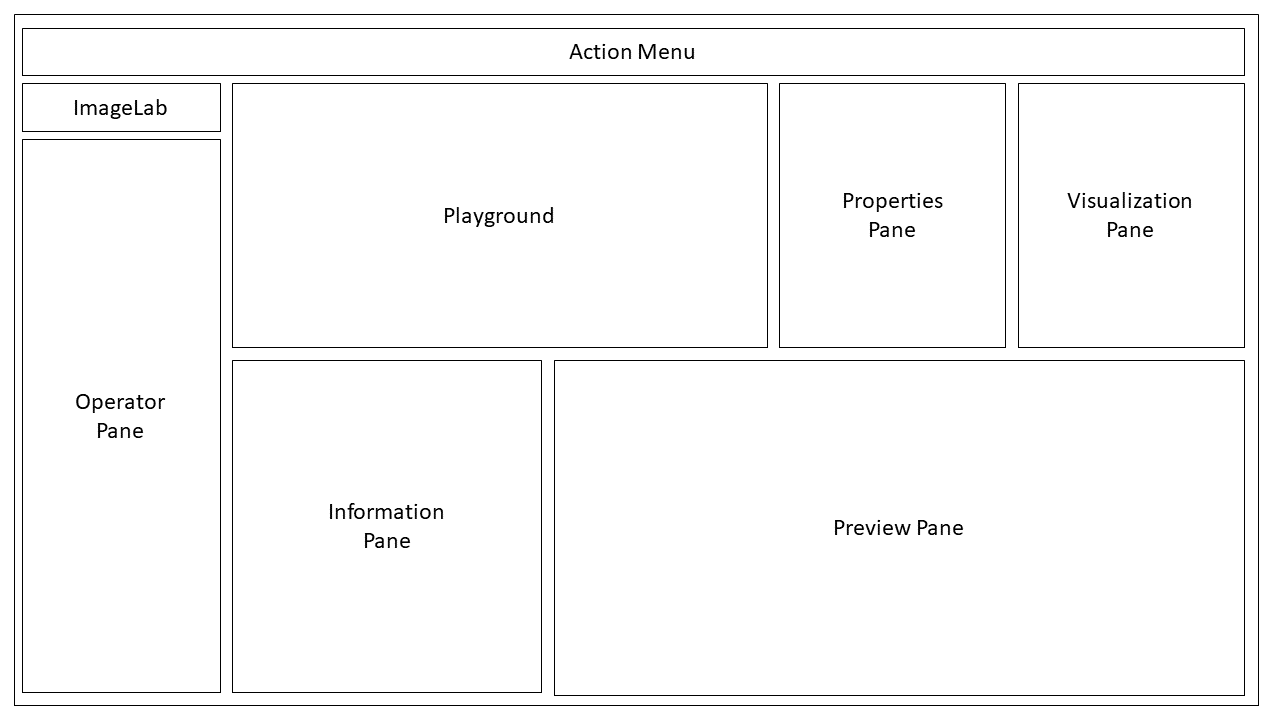}
    \includegraphics[width=.80\textwidth]{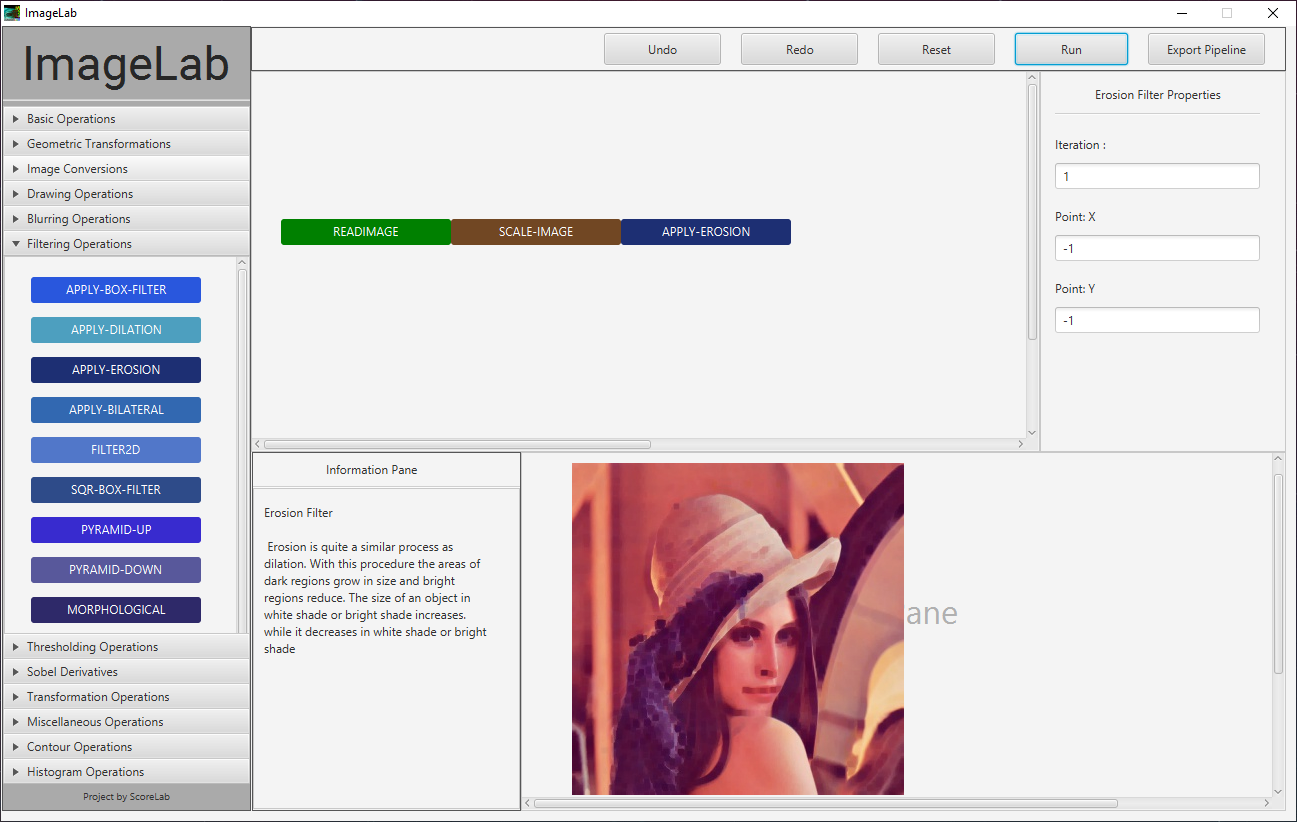}\par
    \caption{Mockup of Imagelab UI with different panels to cater for the different UI requirements with implemented(before usability experiment) initial Imagelab java interface}
    \label{fig:mockup_imagelab}
\end{figure}

ImageLab's interface, illustrated in figure \ref{fig:mockup_imagelab}, is segmented into multiple panels, each with a distinct role to enhance usability. The topmost action menu is equipped with essential functionalities such as execution commands for the image pipeline, an undo/redo feature to modify pipeline changes, and export options for saving current operations. Adjacent to this, the operator panel presents image operations as draggable \textbf{blocks}. This block-driven approach simplifies image operations, catering to both novices and experts. As users drag blocks into the central playground area, they can assemble various pipeline structures for different tasks. Upon placing a block, an associated properties and information pane emerges, providing context about the selected operator along with sample usage guidelines. Activating the "run" command reveals the processed image in a dedicated preview pane, which allows users to witness their adjustments in real-time. Some specific operations, like the histogram operator, also generate visual feedback in a separate visualization pane, enriching the user's understanding of the image's underlying pixel data. This layout not only streamlines the image processing workflow but also minimizes the need for traditional coding. Furthermore, the dynamic feedback provided by the preview pane and the comprehensive range of image operators (detailed in subsequent sections) collectively contribute to a flattened learning curve, making ImageLab an accessible tool for users of varying expertise.

\begin{figure}
    \centering
    \includegraphics[width=.85\textwidth]{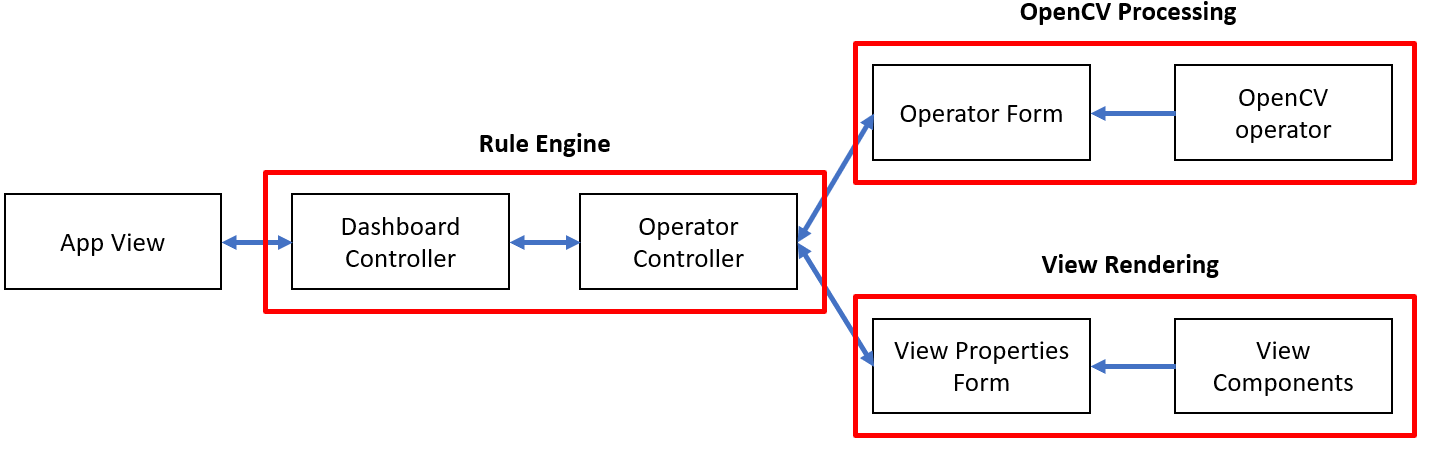}\hfill
    \caption{ImageLab backend architecture to cater the requirements of the UI components}
    \label{fig:flow_imagelab}
\end{figure}

ImageLab's internal system design leverages a pipeline mechanism for image processing, employing a drag-and-drop interface to enhance usability and enable the incremental evolution of the image processing pipeline using blocks. The resulting output is produced by sequentially applying each block. Built atop OpenCV \cite{opencv_library} and Java 8, the application's high-level design is realized through its subdivision into several pivotal components, each intended to maximize application performance and ensure scalability for future feature integration. Key among these are the \textbf{Rule Engine}, \textbf{OpenCV Processing}, and \textbf{View Rendering} components, as illustrated in Figure \ref{fig:flow_imagelab}. The operational flow is such that, given the view is in State $S_{i}$—representative of a specific operator being introduced to the pipeline—the system first verifies rule compatibility with the operator, then engages OpenCV's image processing functionalities, and finally renders the resultant image in State $S_{i+1}$. The continuous addition of operators effectively advances the system's state. To chronicle this state progression, the application incorporates a stack data structure. Beyond merely preserving state history, this stack facilitates seamless undo/redo operations and permits the saving of the pipeline both as a template and as a historical record. This empowers end-users to reference or reuse specific pipelines in future endeavors. Subsequent sections delve deeper into the objectives and functionalities of each component.

The \textit{Rule Engine}, as the moniker suggests, is chiefly responsible for validating the sequence of pipeline operators. The question arises: why is sequence validation pivotal? Given ImageLab's primary goal of imparting image processing knowledge to novices, it is imperative that foundational concepts are presented accurately. Thus, the rule engine embodies this commitment by intercepting any mal-sequenced operator applications. For instance, a sequence like \textbf{READ IMAGE -> SCALE IMAGE by 0.5} is sanctioned, whereas the inverse \textbf{SCALE IMAGE by 0.5 -> READ IMAGE} is disallowed, since scaling necessitates a pre-existing image in the pipeline. Analogously, consecutively stacking identical operators is precluded. These stipulations are meticulously codified in the rule set, ensuring holistic validation. At this component layer, operations manifest as \textbf{BLOCKS}, facilitating users to drag and drop operators effortlessly.

\begin{figure}
    \centering
    \includegraphics[width=.85\textwidth]{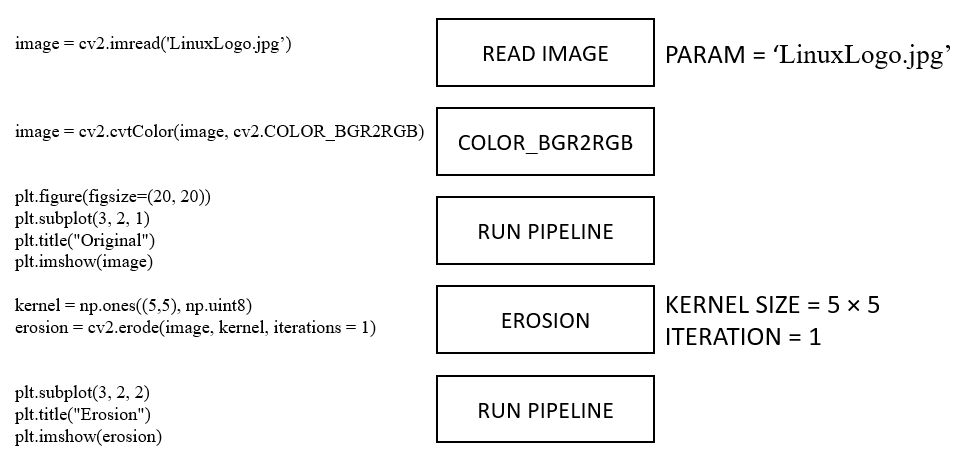}\hfill
    \caption{How Imagelab replace the codes with blocks}
    \label{fig:processing_imagelab}
\end{figure}

In ImageLab, each sequence of blocks represents a series of operations that must be processed to achieve the desired pipeline outcome. The heavy lifting for this processing is facilitated by OpenCV, a robust image-processing library. While OpenCV's native development is in C++, it also supports bindings for various languages, including Java. ImageLab capitalizes on the Java wrapper provided by OpenCV (https://docs.opencv.org/4.x/javadoc/index.html) to tap into its rich feature set for image processing tasks. Predominantly, operations within ImageLab are constructed using OpenCV's \textit{core} and \textit{imgproc} modules. The \textit{core} module aids in loading, storing, and displaying image containers in a matrix format—a prerequisite for numerous image processing effects. It also offers foundational operations like Image Masking and multi-image blending. On the other hand, the \textit{imgproc} module supplements the \textit{core} functionalities, introducing advanced features. Broadly, the OpenCV processing module translates the block sequences validated by the rule engine into tangible image-processing steps by invoking corresponding OpenCV functions, as illustrated in Figure \ref{fig:processing_imagelab}.

The rendering component of ImageLab's architecture is pivotal in ensuring that images are processed according to the operator sequence and subsequently displayed within the user interface. If the processed image dimensions exceed the display area, the pane seamlessly adapts to provide scrolling capabilities, ensuring full image visibility. A significant facet of this component is its integration with the rule engine's stack data structure. This synergy allows the rendering component to chronicle the image's transformation journey. Consequently, users can trace the image's evolution through various pipeline stages, gaining insight into the transformational steps culminating in the final output.

\section{Features and Functionality}

\subsection{Variety of image operator options}

Building upon the integration of the OpenCV module,  harnesses a rich suite of image processing features. The decision to embed OpenCV was driven by its comprehensive capabilities, allowing  to span functionalities from basic I/O tasks to intricate object detection operations, thus encapsulating the domain's salient concepts. Key motivations for selecting OpenCV included its robust community support, seamless Java integration, and cross-platform compatibility. Consequently, ImageLab emerges as a potent \textbf{All-in-one image-processing learning tool}, encompassing features like image I/O, geometric transformations, image conversions, drawing, blurring, filtering, thresholding, Sobel derivations, edge detections such as the canny technique, image segmentation using Laplacian and distance transforms, contour operations, and histogram operations.

\begin{figure}
    \centering
    \includegraphics[width=.85\textwidth]{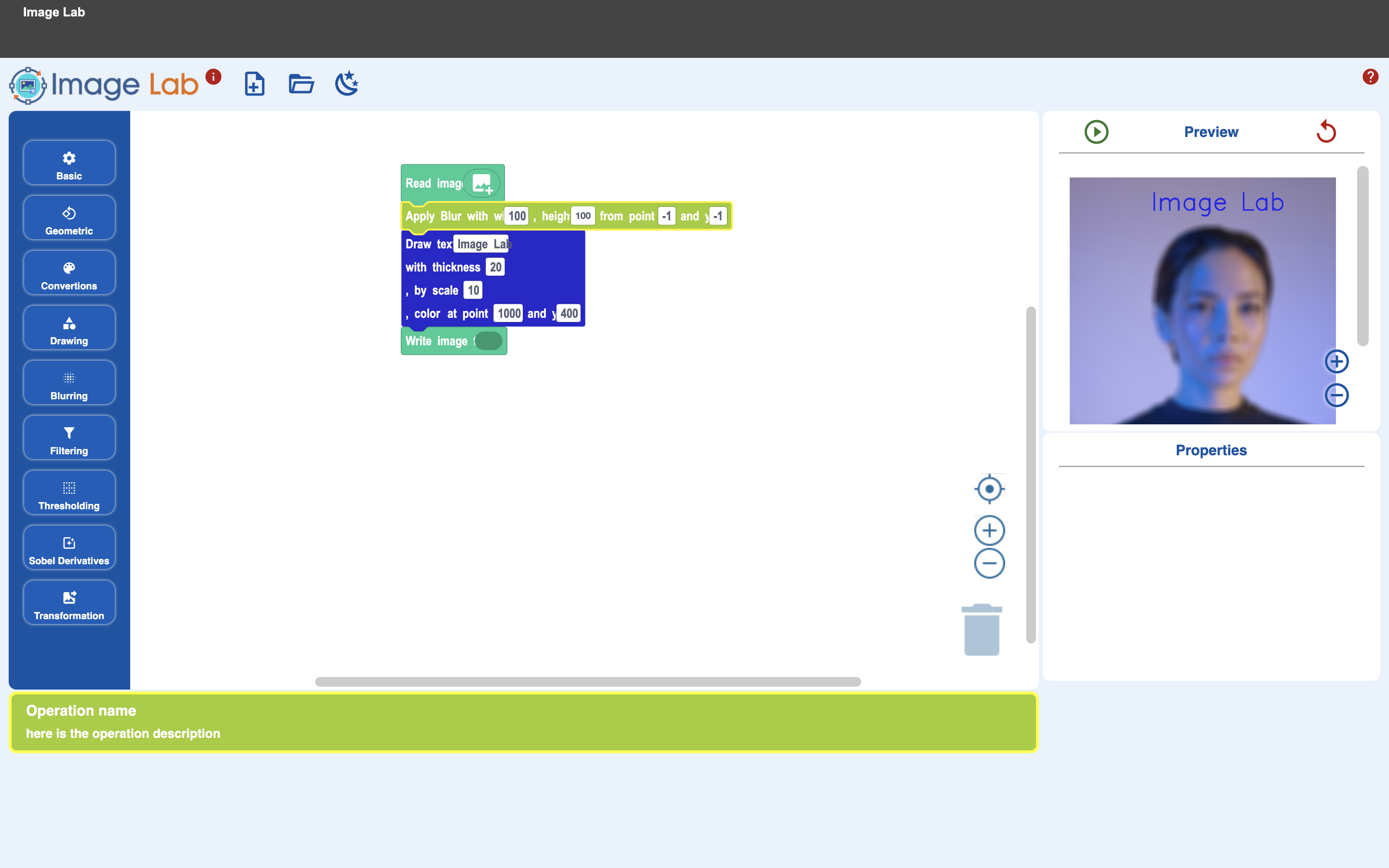}\hfill
    \caption{How Researchers Improve the Imagelab UI with More User-friendly to age-group of Children with Javascript}
    \label{fig:imagelab_elctron}
\end{figure}

\subsection{Easy access with blocks}

 departs from traditional programming practices by introducing the concept of \textbf{blocks}, effectively replacing micro-level code snippets. Such an approach offers several benefits, including accelerated development of intricate image processing pipelines and enhanced clarity for newcomers about which operator combinations yield specific outputs. The cross-platform support of OpenCV, coupled with the Blockly library (\url{https://developers.google.com/blockly}), facilitates the creation of a more intuitive UI using Javascript. For optimized UI performance, the backend, originally written in Java, was transitioned to NodeJS. Although such rewrites can be challenging, the availability of an OpenCV JavaScript plugin simplified the process, requiring mostly core code translation. This approach mirrors the design philosophy of Scratch (\url{https://scratch.mit.edu/}) by MIT, breaking down the \textbf{syntax barrier for learning} and making the tool approachable even for children eager to delve into image processing concepts. As illustrated in Figure \ref{fig:imagelab_elctron}, the transition from a Java UI to a JavaScript integration significantly enhanced performance. Advanced users have the flexibility to choose between the two interfaces, promoting recognition over recall and streamlining the process of selecting the appropriate operator block for the task at hand.

\subsection{Drag and Drop Interface}

The utilization of blocks in encapsulates individual image processing tasks, while its drag-and-drop feature enhances the application's flexibility and user-friendliness. This design fosters a \textbf{flow of learning}, transforming the entire image-processing experience into an intuitive flowchart. The subsequent operators available to users are indicated by puzzle-shaped connectors, allowing beginners to readily discern the logical progression of blocks for their pipeline. This drag-and-drop mechanism not only \textbf{gamifies} the user interface, fostering an exploratory and playful environment, but it also led the researchers to aptly term it the \textbf{Playground}, as opposed to a mere pipeline pane. From a development perspective, while creating such an interactive interface presented challenges in Java, especially when paired with a rule engine backend, transitioning to a JavaScript UI simplified the implementation and augmented it with distinct block types. This transition enables users to navigate the application at their own pace more effectively.

\subsection{Real Time Visual Changes}

 incorporates a feature ensuring that users can observe each micro-level state change in an image. By providing a preview pane, users witness real-time modifications to the image upon applying various operators. The significance of this feature is reinforced by our user study, detailed in the subsequent section, which gathered positive feedback from both novice users, including children, and intermediate users. Such real-time feedback aligns with usability heuristics, ensuring that the system conveys meaningful information promptly.

\section{User Study}

In our endeavour to understand how ImageLab can serve a diverse audience in their journey to learn and delve into image processing, we embarked on a comprehensive usability study. The primary aim of our user study was to gauge the effectiveness and usability of ImageLab as an image enhancement tool when juxtaposed against the conventional Python programming approach using OpenCV. This study was structured into three distinct phases: an initial questionnaire, a user think-aloud session, and a concluding questionnaire.

\subsection{Experimental setup}

The preliminary questionnaire primarily aimed to capture the demographic distribution of our participants. More critically, it sought to identify the essential features that users expect in an image processing tool. From our analysis, we distilled seven core usability features, which we hypothesized would enhance the user experience. Subsequently, during the interactive phase, participants were encouraged to engage directly with ImageLab and its features. To foster a genuine and spontaneous reaction, we implemented a 'think-aloud' approach, enabling users to vocalize their thoughts, concerns, and feedback as they navigated through the tool.

\subsubsection{Participants}
The user study involved a distinctive group comprising school children and first-year university students. Within this collective, the age distribution showcased a blend of participants: 11 individuals fell within the 21 - 25 years bracket, 2 were between 10 - 15 years, and another 2 ranged from 16 - 20 years. This mix not only encompassed younger students potentially new to image processing but also older ones who might have had some exposure to the domain.

Regarding gender diversity, the group consisted of 10 females and 5 males. This balance allowed for an inclusive assessment, capturing insights and experiences from both genders, which is crucial for a well-rounded evaluation of usability.

Lastly, when considering experience levels in image processing or similar tasks, participants' self-assessments were as follows: 8 identified as beginners, 5 as intermediates, and a minority of 2 classified themselves as novices. This tiered representation of expertise provided a multi-layered perspective on ImageLab's usability, ensuring feedback was sourced from users of varying familiarity with the subject. Overall, the broad spectrum of participants in terms of age, gender, and experience level promised a thorough understanding of the ImageLab tool across diverse user profiles.

\subsubsection{Procedure}

\begin{figure}
    \centering
    \includegraphics[width=.35\textwidth]{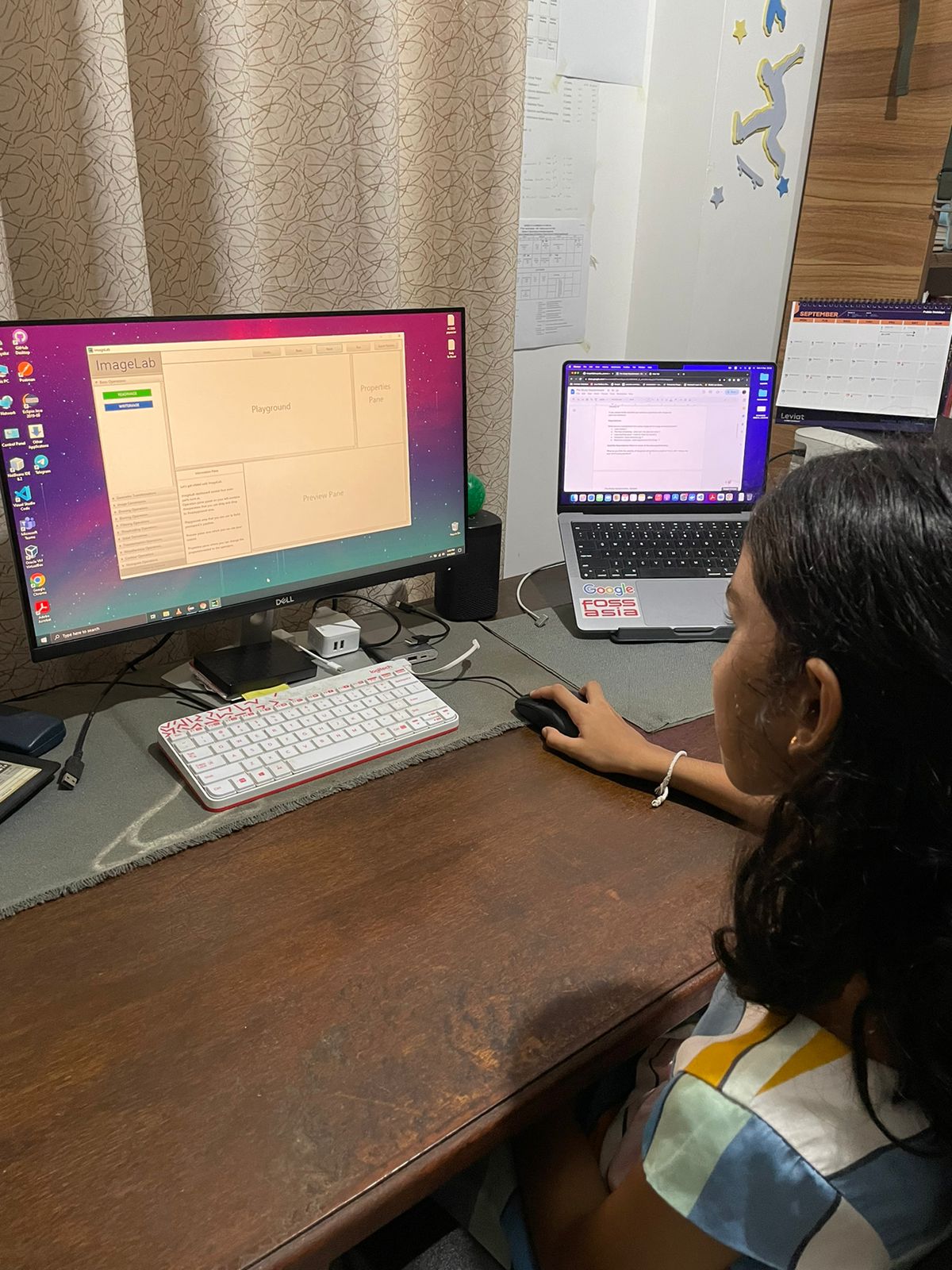}
    \includegraphics[width=.35\textwidth]{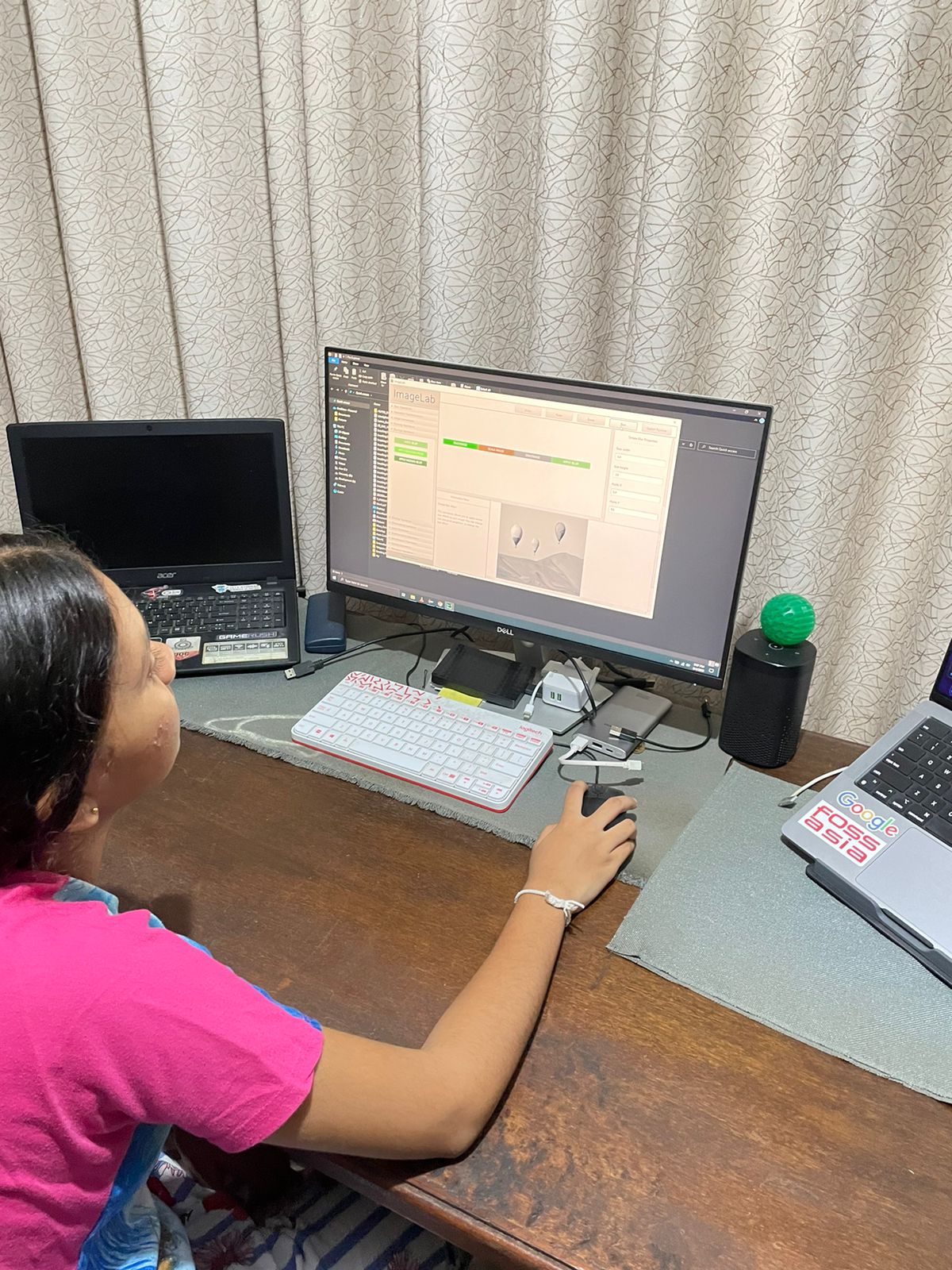}\par
    \caption{School Participants interact with the Imagelab for the experiment}
    \label{fig:imagelab_user_experiment}
\end{figure}

The user study's experimental procedure encompassed two distinct tasks designed to gauge participants' experience with image enhancement using both the Python programming approach and ImageLab.

\textbf{Task 1:} Participants were first presented with an image and tasked with both dilating and eroding it. Their subsequent objective was to compare the enhanced image with the original. To facilitate this, a code template incorporating necessary OpenCV functions was provided. The success metrics for this task included the time participants took to achieve the desired outcome, the complexity of the code they wrote or modified, and a qualitative assessment of the resultant image's enhancement quality.

\textbf{Task 2:} Here, participants were tasked with blurring the given image, first using a 3x3 kernel and subsequently with a 7x7 kernel. As with Task 1, a code template was furnished to aid participants in their endeavor. The same performance indicators from the first task were used to measure the effectiveness and efficiency of their efforts.

For each task, participants were initially directed to employ Python for the image enhancement. Upon completion, they were then asked to replicate the tasks in ImageLab, allowing for a side-by-side comparison of their experiences with both platforms. This structured approach ensured that participants had firsthand exposure to the intricacies and usability of both methods, laying the groundwork for an informed feedback session.

\subsubsection{Data Collection}

Our data collection methodology was bifurcated to capture both the preliminary expectations and post-experiment reflections of the participants, ensuring a comprehensive understanding of their entire journey during the user study.

\textbf{1. Pre-study Questionnaire:} Before commencing the tasks, participants were provided with a "pre-study questionnaire." This initial survey was designed to gauge their prior knowledge, expectations, and initial perceptions about image processing and the tools they would be using. By capturing this baseline data, we aimed to contextualize the subsequent feedback and understand any shifts in perceptions post-task.

\textbf{2. Think-aloud Phase Recording:} During the interactive "think-aloud" phase of the experiment, participants' interactions and vocalized thoughts were recorded. This method aimed to capture their real-time experiences, challenges, and feedback as they navigated through the tasks using both Python and ImageLab. By recording these sessions, we could revisit and analyze their instantaneous reactions, ensuring that no nuanced feedback was overlooked.

\textbf{3. Post-study Questionnaire:} Upon task completion, participants were presented with a "post-study questionnaire." This survey sought to collect their reflections on the usability, efficiency, and overall experience of the tools they interacted with. Comparing this with the pre-study responses allowed us to pinpoint areas of change in their perceptions, insights into the user-friendliness of the tools, and suggestions for improvement.

The combination of structured questionnaires and recorded think-aloud sessions ensured a holistic data collection approach, capturing both quantitative metrics and qualitative insights.

\subsubsection{Data Analysis}

This section elucidates the insights gleaned from the data obtained through the pre-study and post-study questionnaires.

\textbf{1. Pre-study Feedback on Coding with OpenCV:} A predominant sentiment among participants was the iterative nature of coding with OpenCV, where 5 participants felt the need to make multiple attempts for achieving correct results. A combined total of 6 users commented on the extensive nature of coding or felt there was a lack of an immediate change display tool. On the other hand, 6 respondents voiced concerns about the convoluted nature of understanding the best programming approach and discerning the operators to be applied subsequently. The challenges also extended to the initial setup environment, with 2 participants encountering errors.

\textbf{2. Usability Expectations:} Evidently, the desire for "Less coding" led the chart with 12 participants highlighting it, followed closely by the wish for real-time updates on image alterations and a seamless learning trajectory, each endorsed by 10 participants. Interestingly, an equal number of users (7) showed an inclination towards an interactive setup with suggestions, shortcuts, outputs, and progress tracking. The demand for a comprehensive feature-set for image processing experimentation was voiced by 6 users.

\textbf{3. Post-study Feedback:} Following the experiment, participants were asked to evaluate their experience with ImageLab in comparison to their prior understanding of image processing. The distribution of their self-assessed expertise levels showed a majority, 4 participants, identifying as `Intermediate'. Three classified themselves as `Beginners', with one participant each as an `Expert' and `Novice'.

Furthermore, when inquired about the incorporation of `less coding/ease of use' in ImageLab, there was a unanimous consensus, with all 9 participants affirming that this feature was evident and effective in the tool.

To gain a deeper understanding of participants' experiences with ImageLab, they were asked to rate several of its features. The aggregated results are summarized in the table below:

\begin{table}[h!]
\centering
\begin{tabular}{|l|c|c|c|c|}
\hline
\textbf{Feature Evaluated} & \textbf{Mean} & \textbf{Median} & \textbf{Mode} & \textbf{Standard Deviation} \\
\hline
Learning Flow of ImageLab & 3.67 & 4.00 & 3 & 1.00 \\
\hline
Learning Curve of ImageLab & 4.22 & 4.00 & 5 & 0.83 \\
\hline
Interactivity of ImageLab & 3.89 & 4.00 & 4 & 1.05 \\
\hline
Real-time Change Display of ImageLab & 4.00 & 5.00 & 5 & 1.32 \\
\hline
Availability of Operators in ImageLab & 4.33 & 5.00 & 5 & 0.87 \\
\hline
Overall Usability of ImageLab & 4.56 & 5.00 & 5 & 0.73 \\
\hline
\end{tabular}
\caption{Participants' ratings on various ImageLab features.}
\label{table:feedback}
\end{table}

The table indicates a strong inclination towards the user-friendliness, interactivity, and comprehensive features of ImageLab. Participants particularly appreciated the real-time change display and the range of operators available for experimentation. The overall positive ratings underscore ImageLab's potential to revolutionize the image processing experience for its users.

\section{Discussion}

\subsection{Reflections made through the user study}

Throughout the paper, it becomes evident that a systematic approach was employed to establish as a versatile tool suitable for both novice and expert users. The research journey commenced with a survey to ascertain expectations from an image processing tool, followed by a think-aloud experiment with a focus group, and culminated in rigorous data analysis. The findings revealed that successfully met a majority of the usability expectations, with five out of the seven criteria scoring above 4 on a 5-point scale. Beyond mere usability, 's feature set positions it as a comprehensive platform for learning and experimentation in image processing. This is attributed to the study's adept understanding of user needs and its capability to address them. In contrast to existing literature, to the best of the researcher's knowledge, no other study has offered such a holistic feature set tailored to a diverse audience. This research bridges that gap both in terms of usability and the enriched UI/UX features incorporated within the system.

\subsection{Remove the barriers for Children and Novice users to learn image processing}

The user study underscored the ability of children to enhance their understanding by exploring and questioning concepts independently, indicating 's potential as a \textbf{self-learning} platform. This autonomy not only renders learning more engaging but also makes intricate subjects like image processing more digestible. Several school participants expressed heightened interest in the application, often requesting extended interaction time. Additionally, not just children but also novice university students, including those with minimal exposure to image processing, found the tool approachable and intuitive. Such inclusivity effectively diminishes the learning barriers for young and novice enthusiasts, underscoring 's success in obtaining impressive usability scores.

\subsection{Empowerment of Education}

 not only provides children with an exploratory platform for image processing but also nurtures their creativity and critical thinking. By motivating children to think independently, rather than merely replicating content, the tool fosters cognitive growth and enhanced problem-solving capacities. Feedback from parents of the participating children affirmed that has been conscientiously designed, keeping in mind both age-appropriate content and security considerations. Thus, parents and educators can confidently introduce children to, assured of its educational merit and safety.

\subsection{Comprehensive Tool for Learning}

Researchers observed that ImageLab serves as a pivotal tool for university students specializing in image processing or related domains, thanks to its extensive features powered by OpenCV. This platform facilitates hands-on learning and experimentation, catering to both novices and experts alike. Feedback from university participants highlighted ImageLab as an invaluable resource for both research projects and coursework. Its expansive feature set efficiently streamlines tasks, promoting rigorous experimentation. Furthermore, researchers envision that ImageLab's utility extends beyond academic pursuits. Students equipped with ImageLab will be better poised for professional roles in image processing, offering practical expertise across various industries and enabling them to undertake in-app projects.

\section{Future Work}

While ImageLab already shows promise, there are numerous avenues for its further augmentation. One notable improvement is the introduction of operator suggestions post their drag-and-drop onto the playground, allowing users the discretion to heed or disregard these prompts. Additionally, integrating pre-defined templates for commonly used image processing tasks, like the canny edge detection pipeline, could be beneficial. Venturing into contemporary computer vision domains, ImageLab might incorporate advancements like embedding deep learning models and facilitating video analysis. A particularly impactful enhancement would be the ability to concurrently process multiple images, enabling both experts and novices to juxtapose various pipelines in real-time.

\section{Conclusion}

Researchers have a strong belief that this tool can be ideally matched with people who are keen on learning, discovering, and putting foundational knowledge on image processing. Also, many universities and high educational institutions can use this tool as a guiding tool to explain the complex areas of image processing in a very easy manner as a visual tool. Imagelab's usability and improved UI/UX have the potential to transform the educational landscape for both children and university students. It provides a user-friendly and educational platform for children to explore image-processing concepts, fostering their creativity and problem-solving skills. Imagelab's future work envisions a more user-centric and efficient tool as Imagelab continues to evolve, these enhancements represent an exciting step toward democratizing image processing and promoting accessible, engaging, and educational interactions with this powerful technology.

\bibliographystyle{unsrt}
\bibliography{main}

\end{document}